\newcommand{\dino}{DINOv2\xspace}
\newcommand{\clip}{CLIP\xspace}
\newcommand{\sam}{SAM\xspace}
\newcommand{\z}{\mathbf{z}}
\newcommand{\tx}{\mathbf{t}}
\newcommand{\x}{\mathbf{x}}
\newcommand{\pt}{\mathbf{t}\xspace}
\newcommand{\im}{\mathbf{im}\xspace}

\newcommand{\ESupCon}{\text{TSupCon}\xspace}
\newcommand{\ours}{\text{FMbSeg}\xspace}

\newcommand{\eg}{e.g.\@\xspace}

\newcommand{\etc}{%
    \@ifnextchar{.}%
        {etc}%
        {etc.\@\xspace}%
}
\newcommand{\myparagraph}[1]{\noindent\textbf{#1}}
\newcommand{\cls}{\mathbf{cls}\xspace}
\newcommand{\suppl}{Appendix\xspace}
\documentclass[10pt,twocolumn,letterpaper]{article}

\usepackage[pagenumbers]{wacv} 

\usepackage{graphicx}
\usepackage{amsmath}
\usepackage{amssymb}
\usepackage{booktabs}

\usepackage{xcolor}
\usepackage{algorithm}
\usepackage{algorithmic}
\usepackage{multirow}
\usepackage{array}
\usepackage{pifont}
\usepackage{arydshln}

\usepackage[pagebackref,breaklinks,colorlinks]{hyperref}

\usepackage[capitalize]{cleveref}
\crefname{section}{Sec.}{Secs.}
\Crefname{section}{Section}{Sections}
\Crefname{table}{Table}{Tables}
\crefname{table}{Tab.}{Tabs.}

\def\wacvPaperID{2677} 
\def\confName{WACV}
\def\confYear{2025}

\begin{document}

\title{Annotation-Free Semantic Segmentation with Vision Foundation Models}

\author{Soroush Seifi*
\and 
Daniel Olmeda Reino
\and
Fabien Despinoy
\and
Rahaf Aljundi
\and
Toyota Motor Europe (*contracted services)\\
{\tt\small firstname.lastname@toyota-europe.com}
}

\maketitle

\begin{abstract}
\vspace{-0.4cm}
Semantic Segmentation is one of the most challenging vision tasks, usually requiring large amounts of training data with expensive pixel level annotations. With the success of foundation models and especially vision-language models, recent works attempt to achieve zeroshot semantic segmentation while requiring either large-scale training or additional image/pixel level annotations. In this work, we generate free annotations for any semantic segmentation dataset using existing foundation models. We use CLIP to detect objects and SAM to generate high quality object masks. Next, we build a lightweight module on top of a self-supervised vision encoder, DinoV2, to align the patch features with a pretrained text encoder for zeroshot semantic segmentation. Our approach can bring language-based semantics to any pretrained vision encoder with minimal training, uses foundation models as the sole source of supervision and generalizes from little training data with no annotation.
 \end{abstract}
\vspace{-0.5cm}
\section{Introduction}
\label{sec:intro}
\vspace{-0.2cm}
Large-scale and inexpensive training data has recently enabled the surge of foundation models in computer vision~\cite{bommasani2021opportunities}. These models have been employed to avoid expensive annotations and computational requirements of many vision tasks~\cite{radford2021learning,kirillov2023segment, zhang2022dino, oquab2023dinov2,alayrac2022flamingo}. Leveraging foundation models for semantic segmentation requires the model to produce pixel-wise predictions on new datasets, domains and ontologies. Therefore, the model must be \textbf{1)} promptable with an open set of categories and \textbf{2)} highly discriminative for dense recognition tasks. Consequently, deploying foundation models to obtain cheap annotation for semantic segmentation is challenging as existing models lack either semantic awareness~\cite{zhang2022dino,liu2022convnet, kirillov2023segment} or local feature robustness~\cite{radford2021learning,jia2021scaling}.

In this paper, we propose a novel approach to open vocabulary semantic segmentation by composition of different foundations models as building blocks and source of supervision. In particular, we employ Contrastive Language-Image Pretraining (CLIP~\cite{radford2021learning}), trained on a large set of image-text pairs, to derive a semantic understanding of different regions within an image \cite{zhou2022extract,oquab2023dinov2}. We use Segment Anything (SAM~\cite{kirillov2023segment}), to supervise the accurate shape and extension of objects detected by CLIP.
Next, we train a lightweight module that aligns the generic visual features of a self-supervised task-agnostic foundation model, DinoV2, with the text embedding
space of a CLIP model in a contrastive manner. The result is a model that is highly discriminant, generalizable and grounded to semantic meaning and object shape without requiring human-generated segmentation or image annotations.

\begin{figure}[t]
    \centering
   {\includegraphics[width=\linewidth]{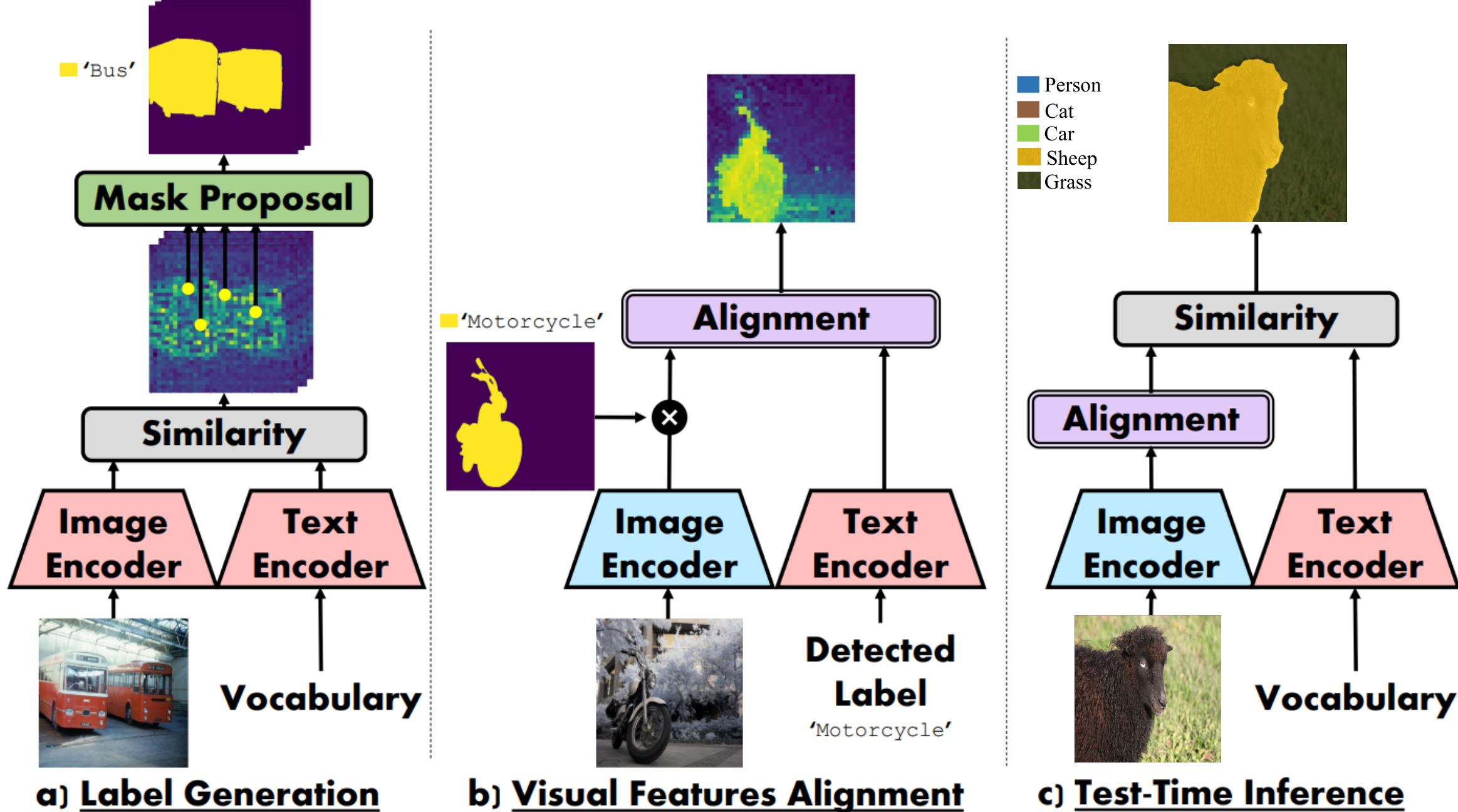}}
   \vspace{-0.5cm}
    \caption{\small \textbf{Overview of our method (\ours)}.  a) \textbf{Label Generation:} We detect object and background categories in an image using a frozen pretrained image-text model (\eg \clip, in pink). We select image patches with high similarity to the text representation of the detected categories. We pass the location of those patches to a mask proposal network (\eg SAM, in green). b) \textbf{Visual Features Alignment:} We use the generated segmentations and detected categories to align features from a more expressive frozen image encoder (e.g. DINOv2, in blue) with a frozen pretrained text encoder. c) \textbf{Test-Time Inference:} At test time, the newly aligned image encoder projects image features into text space. Every pixel is classified according to their similarity to the pre-computed text prototypes of a target ontology.}
    \vspace{-0.5cm}
    \label{fig:teaser}
\end{figure}

An overview of our method coined as \ours is shown in Fig.~\ref{fig:teaser}.
The main contributions of this paper are:  (1) We propose a method to generate  semantic segmentation pseudo annotations with zero pixel or image level labels using CLIP and SAM. (2) We propose composing pretrained language-image and self-supervised vision foundation models by means of a lightweight contrastive alignment module trained with a uniquely designed loss function. (3) We apply the composed model to the zeroshot semantic segmentation setting, where we show generalization to never-seen-before semantic segmentation datasets. (4) We achieve state of the art results on annotation-free semantic segmentation task. 

\vspace{-0.4cm}
\section{Related Work}
\vspace{-0.2cm}
\label{sec:relatedwork}
\paragraph{Vision Foundation Models}
Recent advancements in large-scale pretraining have led to powerful generalist models, transferable to various tasks and domains ~\cite{bommasani2021opportunities}. 
In particular, expressive vision-language models emerge when scaling contrastive pretraining on text-image pairs, as shown by ALIGN and CLIP~\cite{radford2021learning,jia2021scaling}. These models have  been since applied  to a range of vision problems, including open vocabulary semantic segmentation~\cite{Ding_2022_CVPR,xu2022groupvit} and image generation~\cite{ramesh2021zero}. 
SAM~\cite{kirillov2023segment} is a powerful image segmentation model, trained iteratively with weak annotations on one billion images. SAM is not grounded to semantic meaning and although there was an initial attempt by the authors to align SAM with CLIP text embeddings, no quantitative results were reported. 
DINO~\cite{caron2021emerging} and DINOv2~\cite{oquab2023dinov2} are self-supervised image encoders demonstrating high performance in dense prediction tasks. However, DINO models are not aligned with text, making their application to Open Vocabulary scenarios not straightforward. In this work we show an example of foundation models compositionality, where knowledge from different models (CLIP~\cite{radford2021learning}, SAM~\cite{kirillov2023segment} and DINOv2~\cite{oquab2023dinov2}) are incorporated to build an open vocabulary semantic segmentation model.
\\
\myparagraph{Open vocabulary semantic segmentation.}
Zeroshot semantic segmentation requires a model to provide pixel level labels for unseen categories~\cite{bucher2019zero}. Open vocabulary semantic segmentation generalizes this problem aiming at segmenting an arbitrary set of classes. Zhao et al.~\cite{zhao2017open} first introduced the open vocabulary setting by learning a joint embedding for pixels and textual concepts based on WordNet.\\ 
\myparagraph{CLIP-based methods.}
The general trend for open vocabulary methods is to build on top of CLIP for better generalization with less supervision~\cite{li2022grounded,zhang2022glipv2,chen2023open,Ding_2022_CVPR, xu2023masqclip,zhong2022regionclip,rao2022denseclip}. LSeg\cite{li2022languagedriven} matches the pixel embeddings to text embedding of CLIP. OpenSeg~\cite{ghiasi2022scaling} learns visual-semantic alignments by aligning each word in a caption to one or few predicted masks. A popular line of work 
relies on training a class-agnostic mask proposal network, leveraging \clip embeddings for mask classification with additional techniques to strengthen \clip image patch embeddings~\cite{ xu2022simple,xu2023masqclip,han2023open, ding2023open,yu2023convolutions}. These works still require  various degrees of pixel-level supervision, which can be expensive to obtain for fine-grained categories.
\\
\myparagraph{Training-free methods.}
ReCo~\cite{shin2022reco} builds a dataset of $K$ images for each concept from a large-scale unlabelled dataset. Then a nearest neighbour approach is used to produce initial segmentation of a target concept that is then refined by DenseCLIP~\cite{rao2022denseclip}.
CLIP-DIY~\cite{wysoczanska2024clip} divides an image into smaller patches that are then classified by CLIP. Patches are then aggregated and transformed to dense prediction using objectness scores from pretrained foreground-background segmentation model. In our work, we rely on SAM to provide accurate masks of detected objects. All training free methods, require a large preprocessing time and usually many steps of refinement, rendering them infeasible. We use the training-free part of our method (section \ref{subsec-firstpipeline})  as a source of pseudo annotations to produce an efficient model.
\\
\myparagraph{Training-based methods without pixel level supervision.}
Many methods modify the CLIP encoder and update it based on a large set of image-text annotated pairs. 
GroupVit~\cite{xu2022groupvit} optimizes a hierarchical pixel grouping strategy integrated in a learned ViT model. 
TCL~\cite{cha2023learning} trains a decoder to ground masks with language based on datasets of 12M and 3M images. ZeroSeg \cite{chen2023exploring} distills localized semantic information from multi-scale views to a segmentation model via different loss functions. Closer to our work, MaskCLIP \cite{zhou2022extract} utilizes CLIP to generate annotations for training a complete segmentation network from scratch. These works are restricted to CLIP due to its language capabilities and hence require large scale training to create discriminate patch-level features. To the contrary, we map DINOv2~\cite{oquab2023dinov2}  accurate patch features to CLIP text space with minimal training. 
SAM-CLIP~\cite{wang2023sam} brings semantics to SAM by large-scale fine-tuning and distillation from both SAM and CLIP models with 41M unannotated images. Our work only employs SAM as a segmentation teacher and aligns a more expressive vision encoder with CLIP's pretrained language encoder, hence requiring much less training data and no model retraining. 
Particularly, our method surpasses previous works performance using only 118 thousand unlabeled images. Our work is the first to align off-the-shelf vision and text encoders at the patch level with minimal training and a lightweight alignment module, making our method readily accessible for annotation-free semantic segmentation  and usable by practitioners for plug and play semantic segmentation methods on various domains.
\vspace{-0.4cm}
\section{Method}
\vspace{-0.2cm}
\paragraph{General overview.}
In this work, we deploy 3 foundation models by composition for the task of semantic segmentation:
 1) \clip~\cite{radford2021learning}, pretrained on a million image-text pairs, exhibiting semantic understanding of the image as a whole but not designed for object localization. 
 2) \sam \cite{kirillov2023segment}, trained to segment objects or parts of objects, but lacking proper semantic understanding.
 3) \dino~\cite{oquab2023dinov2},
producing features that transfer well to many downstream tasks, and that are consistent across similar objects and parts of objects, but without an explicit link to semantic notions.
We design a method that leverages the distinctive properties of those foundation models to enable zeroshot and open vocabulary semantic segmentation.\\
To achieve this, we propose a two-stage approach:
first, we leverage \clip and \sam to generate pseudo semantic masks for a given vocabulary of classes.
In the second stage, we use the predictions of the first stage to train a small alignment module that aligns a frozen off-the-shelf image encoder, ~\dino~\cite{oquab2023dinov2}, with  a pretrained text encoder at the patch level, resulting in a strong self-supervised semantic segmentation model.
\subsection{Preliminary}
\label{subsec-preliminary}
\vspace{-0.2cm}
We consider a dataset of images $\mathbf{D}=\{im_i\}^M_{i=1}$ accompanied with a set of categories in the vocabulary $V = \{class_1,\dots,class_k,\dots,class_K\}$.
\clip's transformer-based image encoder~\cite{dosovitskiy2020image} takes as input an image ${\im_i} \in\mathbb{R}^{C,H,W}$ divided into patches $\im_i= [im^1_i,\dots im^p_i,\dots,im^N_i]$ and extracts a class token $\cls_i$ and patches embedding $\x_i= [x^1_i,\dots, x^p_i,\dots,x^N_i ]$.
\clip's text encoder takes as input a text description \{a photo of a ``$class_k$''\} and produces the text feature $\tx_k$, with $class_k$ $\in V$ corresponding to the image label. Visual and textual features represented by $\cls_i$ and $\tx_k$ are separately projected to a joint embedding space $\mathbb{R}^D$ and the cosine similarity between them is maximized during \clip's training. While the original CLIP architecture discards the patch features $\x^p_i$, they can be projected onto the same space  $\mathbb{R}^D$. This would enable us to compute the similarity of any category in $V$ with individual patches $\x^p_i$ and produce a rough localization of the objects in the image.
Refer to the \suppl for more details on the specific architecture for projecting patch features $\x^p_i$ to $\mathbb{R}^D$.
\subsection{ Stage 1: Object detection \& masks generation}\label{subsec-firstpipeline}
\vspace{-0.2cm}
In this section, we outline our strategy to generate pseudo semantic segmentation labels using ~\clip and \sam. We employ \clip to recognize categories present in the image and \sam for mask generation. We propose two complementary methods for this, Stage 1.1 and Stage 1.2. We provide further details and examples in the \suppl.

\subsubsection{Stage 1.1: Querying \sam with \clip} \label{subsec-1.1}
\vspace{-0.2cm}

\myparagraph{High-resolution feature extraction} We (over-)sample each image into a high resolution one and divide it into $C$ crops in a sliding window fashion. We then process each crop separately with \clip and rearrange the patches returned from \clip (section \ref{subsec-preliminary}) for all crops to construct the features for the full image. This guarantees precise and high quality feature map $f_i$ for each image. Besides, for each crop $c$, we extract a classification token $\cls_c$. We refer to  the \suppl for details and visualizations.

\myparagraph{Defining the set of concerned categories.}
To detect classes present in an image our method computes the similarity between the classification token for each crop $\cls_c$ and a set of text features $\{\tx_k\}$  corresponding to descriptions extracted from vocabulary $V$ (section \ref{subsec-preliminary}). All possible labels in a given dataset are considered as the vocabulary (e.g. a set of 171 classes for COCO-Stuff).

\noindent\myparagraph{Object presence detection.}
Each crop $c$ is classified with an object category when the object's text feature $\textbf{t}_k$ has the highest similarity to the classification token $\cls_c$ among all descriptions extracted from vocabulary $V$. An object category is considered as present in the image if it has been assigned to more than a predefined number of crops (set to 1 in our experiments).

\noindent\myparagraph{Pseudo mask generation.}
We compute a similarity matrix between the full image feature map  $f_i$ and the text features for the \textbf{detected} categories (Fig.~\ref{fig:alignment_heatmaps}). For each category $k$ we select 5 patches with the highest similarity as query points. We feed these points along with the original image to \sam and select the mask with the highest confidence  $m_i^k$.

\subsubsection{Stage 1.2: \sam masks classification}
\vspace{-0.2cm}
\noindent Stage 1.1 may ignore small objects or generate partial masks for an image (Fig.~\ref{fig:Stag1}) due to sub-optimal query points. Thus we  perform a complementary pseudo label generation mechanism  to further boost the performance.

\noindent{\myparagraph{Automatic mask generation.}
We retrieve all possible masks extracted from the full image using \sam's automatic mask generation pipeline. We constrain the masks by size and predicted IOU to filter out the low quality and duplicate masks (See \suppl).

\noindent\myparagraph{Mask labelling.}
Given the generated high resolution feature map $f_i$ and the detected categories in Stage.1, to classify the generated masks  we compute  mean feature corresponding to the area covered by each mask and compute its similarity to the text features of the \textbf{detected} categories in the image. The class with the highest similarity is selected as the pseudo label for the corresponding mask  $m_i^k$. 

Given the high-resolution feature map $f_i$ and the detected categories from Stage 1, we classify the generated masks by computing the mean feature for each mask’s area. We then compare this mean feature to the text features of the detected categories in the image. The class with the highest similarity $class_k$is chosen as the pseudo label for the corresponding mask  $m_i^k$.
\subsection{Stage 2: Lightweight semantic segmentation}\label{subsec-alignmentmodule}
\vspace{-0.2cm}
Stage 1 extracts segmentation masks from a dataset with a predefined vocabulary, but these masks can be noisy. Additionally, querying two foundation models can be inefficient under low compute constraints. To address this, we propose using the annotations generated in Stage 1 to align any off-the-shelf pretrained vision encoder with text semantics,  with \emph{no human supervision}.\\
\myparagraph{Alignment module.}
We use the generated masks in~\ref{subsec-firstpipeline} as pseudo labels to train a simple alignment module that maps image patch features to text embeddings. This mapping grounds any pretrained vision encoder with language for dense prediction tasks. To avoid biases in supervised models, we focus on self-supervised pretrained models, specifically \dino~\cite{oquab2023dinov2}, which is trained fully self-supervised without text alignment. To handle noisy pseudo annotations, we rely on: 1) frozen pretrained text features as anchors, 2) already discriminative image patch features, and 3) a uniquely designed loss function that is robust to noise.

\myparagraph{Pseudo label assignment.}
Let $\mathbf{D}$ be a dataset of unlabelled images $\mathbf{D}=\{im_i\}^M_{i=1}$. Using the first stage of the pipeline, we extract object masks for each image $im_i$ along with their assigned text features $\{m_i^s,t_s\}$.  
The output of the image encoder (\eg \dino) is $\x_i= [\x^1_i,\dots, \x^p_i,\dots,\x^N_i ]$ where $N$ is the number of patches, ignoring the $\cls$ token. 
 Using the generated masks and their associated detected categories we assign to each patch $\x^p_i$ a pseudo label $y^p_i\in \{1,..,K\}$ where $K$ is the number of detected categories in the dataset $\mathbf{D}$.  
 From the text encoder we extract $K$ text features $T=\{\tx_1,\tx_2,\dots,\tx_K\}$.

 \myparagraph{Training the alignment module.}
The pretrained image encoder remains frozen and we optimize a small alignment module $\mathcal{M}$ to map the patch representations $\{\x^p_i\}$ to the CLIP text embedding space $z^p_i=\mathcal{M} (\x^p_i) \in \mathbb{R}^D$. Note that the method is agnostic to the specific encoder used.
As we strive for simplicity, we design our alignment module as one transformer block with multi-head self-attention layer. The self-attention layer allows each patch to attend to other patches in the image. 
Since we aim for an open-set semantic segmentation, cross-attention with text features~\cite{alayrac2022flamingo,zou2023generalized} is not used as it would require a joint processing of the image features and a closed set of text features at test time.\\
For notation clarity, we drop the image index and consider only an across-batch patch index $i; i \in \{1, \dots,N*B\}$ where N is the number of patches in an image and $B$ is the batch size.
 CLIP~\cite{he2023clip} contrasts the similarity of text features with image class tokens using a cross entropy loss, where a one to one correspondence exists between an image class token and its text features. In our case, we have many image features $\z_i=\mathcal{M}(\x_i)$ extracted from patches of many images and few text features corresponding to detected categories in $\mathbf{D}$. Upon early experiments with CLIP loss, we found it not scalable to image patches and exhibiting poor convergence. \\
We thus propose  to  treat each text feature  $\tx_k$ as a prototype of each  semantic category.
The similarity of a patch feature $z_i$ is to be maximized with the corresponding text prototype $\tx_k; y_i=k$ and with other patches of the same category, from  any image in the same batch.
Note that all the feature vectors (text and image patches) are normalized to have unit norm, and the similarities are expressed as a dot product.

Inspired by supervised contrastive loss SupCon~\cite{khosla2020supervised} that operates on positive and negative pairs, we construct two types of pairs: pairs of image patches (patch-patch pairs: $<\z_i,\z_j>$) and pairs of image patches and text features (patch-text pairs: $<\z_i,\tx_k>$). A patch-patch pair $<\z_i,\z_j>$ is considered positive if the patches belong to the same category class $y_i=y_j$ and negative otherwise. Patch-text pairs $<\z_i,\tx_k>$ are positive if $y_i=k$.
We construct a loss function of two terms operating on the two types of said pairs:
\begin{equation}\label{eq:Esupcon}
\ell_{\ESupCon}=\frac{1}{B*N+K}\left(\sum_{k=1}^{K}  \ell_\pt(\tx_k)  +\sum_{i}^{B*N}\ell_\im (\z_i)\right),
\end{equation}
where B is the batch size and N is the number of  patches in an image; K is the number of text features. 
$\ell_{\pt}$ is  designated for optimizing patch-text pairs 
\begin{equation}
\ell_{\pt}(\tx_k)=\dfrac{1}{N_k}\sum_{i:y_i=k} \ell_{\tx}(\z_i,\tx_{k}),
    \label{eq:l_clSup}
\end{equation}
where $N_k$ is the number of patches with label $y=k$.
\begin{equation}
  \begin{aligned}[b]
&\ell_{\pt}(\z_i,\tx_{y_i}) = -\z_i^\top\tx_{y_i}\\ 
&+ \log\left(\sum_{k=1}^{K}\exp(\z_i^\top\tx_k)+\sum_{j\ne i}\exp(\z_i^\top\z_j)\right).
\end{aligned}
\label{eq:l_clSup_pairwise}
\end{equation}
The loss $\ell_{\pt}(\tx_k)$, defined for each text feature $\tx_k$, considers all the patches that belong to the category $y=k$ , represented by the text feature $\tx_k$. The loss  is minimized by maximizing the similarity of the concerned patch-text pairs, normalized over all other constructed pairs (patch-patch and patch-text pairs) for a given patch $\z_i;y_i=k$ .

The loss  applied to each patch feature  is defined as follows:
\begin{equation}
\ell_\im(\z_i)=\frac{1}{N_{y_i}}\sum_{l:y_l=y_i}\left(-\z_i^\top\z_l + \log \sum_{j\ne i}\exp({\z_i^\top\z_j)}\right),
\label{eq:l_SupCon_pairwise}
\end{equation}
where $N_{y_i}$ is the number of patches in the batch that have the same label as $y_i$. This loss term operates on the image patches and it maximizes the similarities of the patches of the same category across different images. This term is a generalization of SupCon~\cite{khosla2020supervised} to image patches with no data augmentation.

A similar loss function has been proposed for image classification in~\cite{aljundi2023contrastive} and it was shown that $\ell_t$~\eqref{eq:l_clSup_pairwise} formulation  can be expressed as a smooth approximation to the maximum function of SupCon and Cross Entropy (CE) loss; in our case  of SupCon on patch-text pair  and CE on the patch feature with the corresponding text feature as a representative class prototype. This approximation is key to allow a smooth optimization of the different similarities while tolerating possible noisy pairs. 
We optimize the alignment module $\mathcal{M}$ with $\ell_{\ESupCon}$ (Eq. ~\ref{eq:Esupcon}), for a fixed number of epochs and  ablate different loss functions in Section~\ref{sec:loss-ablation}.

\myparagraph{Deployment.}
At test time, we extract $\{\x_i\}$ features from the image, using the frozen image encoder. The image patches are forwarded through $\mathcal{M}$, $\z_i=\mathcal{M}(\x_i)$, then we compute for each $z_i$ the most similar text feature $\tx_k$, after which we assign to $\z_i$ a label $k$. Text features are precomputed using a frozen text encoder.

\myparagraph{Pixelwise Segmentation.}
Since our alignment module works at patch-level, to generate pixel level predictions, we interpolate the similarities to the original image dimension, we call this model \ours-Stage 2 (base). 
For more refined and accurate segmentation, we leverage SAM. We label the automatically generated masks by SAM using our alignment module, based on the majority vote of the classified patches within each mask. As SAM might miss certain regions depending on the hyperparameters set for mask quality and size, we overlay the interpolated segmentation with the labeled masks for a complete result. We call this model \ours-Stage 2 (refined). Note that any off-the-shelf segmentation model, such as Efficient-SAM~\cite{xiong2024efficientsam} or Fast SAM~\cite{zhao2023fast}, can be used.

\label{sec-method}

\vspace{-0.3cm}
\section{Experiments}
\vspace{-0.1cm}
\subsection{Experimental Setup}
\vspace{-0.1cm}
\myparagraph{Implementation Details.}
We consider Vit-L/14 for both CLIP and DINOv2. We train our alignment module with  SGD optimizer, cosine annealing scheduler and a batch size of 5 images. We 
use COCO dataset's~\cite{caesar2018coco} unlabelled images for pretraining our alignment model. Note that other training-based zeroshot semantic segmentation methods use either COCO \cite{cha2023learning, he2023clip, chen2023exploring} or a much larger dataset with labels \cite{xu2022groupvit,wang2023sam}  to optimize their model (table \ref{tab:segmentation-datasets}). We  extract pseudo segmentations of COCO-Stuff  vocabulary  using the first stage of our pipeline (section ~\ref{subsec-firstpipeline}). Next, we use them to train our alignment module for 10 epochs only. 
Our results are generated in a much more constrained setting using only COCO-stuff images and class list without any groundtruth annotations.
\begin{figure*}[t!]
    \centering
    \includegraphics[width=\linewidth]{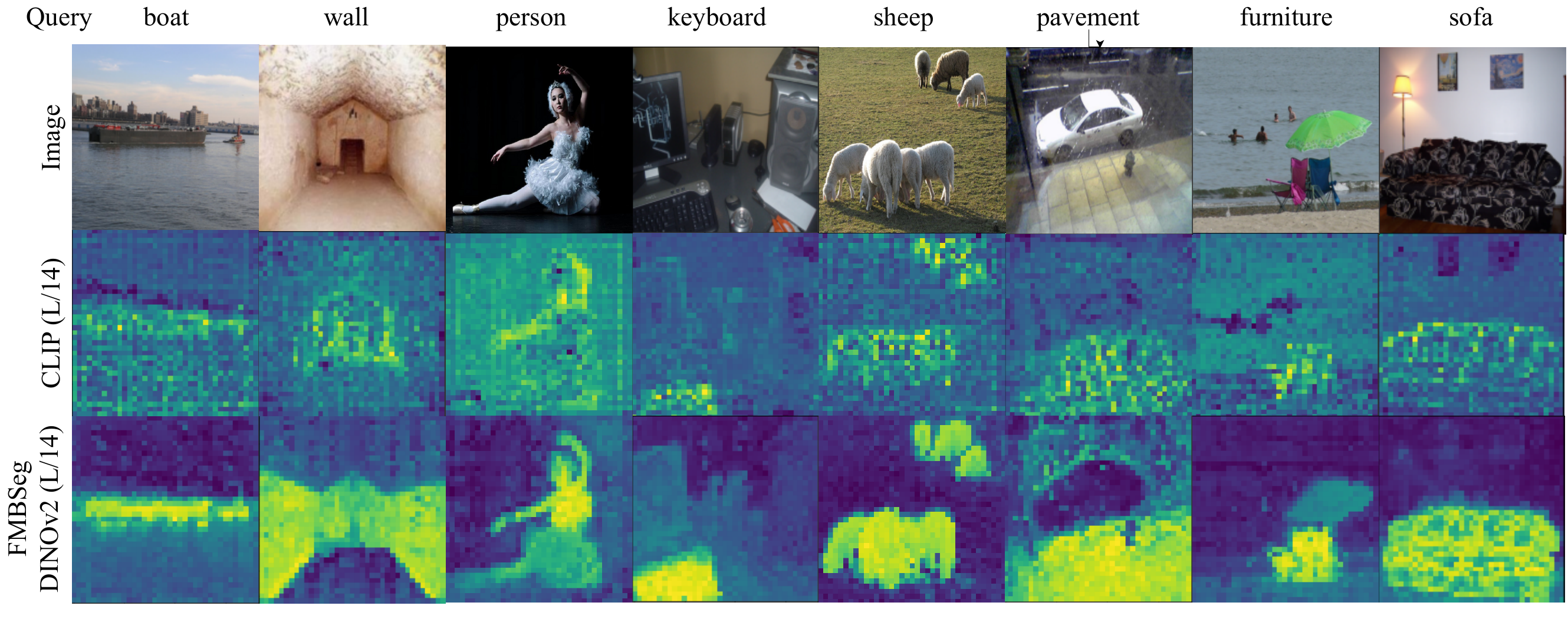}
    \vspace{-0.7cm}
    \caption{\small \textbf{Patch level alignment between image and class}. First row shows images from Pascal VOC. Second row shows the similarity between patch features from CLIP and the text features of the detected category. Third row shows the similarity map after aligning a DINOv2 model with \ours.}
    \vspace{-0.8cm}\label{fig:alignment_heatmaps}
\end{figure*}
\begin{figure*}[t!]
    \centering
    \includegraphics[width=\textwidth]{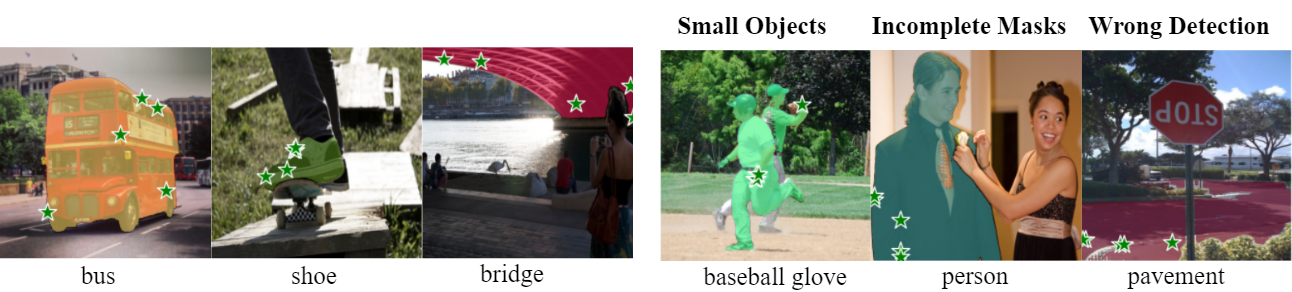}
  \vspace{-0.5cm}
  \caption{ \textbf{Qualitative evaluation of Stage 1.1}. SAM query points generated by our method are shown in green stars. Left shows instances of correct segmentations by Stage 1.1. and Right demonstrates its limitations; small objects, wrongly detected classes (due to ambiguities) and not enough query points to cover all instances. Stage 1.2 alleviates the issue with small objects and incomplete masks since it labels all the masks generated accurately by \sam.}
  \vspace{-0.6cm}
    \label{fig:Stag1}
\end{figure*}
\begin{figure*}
 \vspace{-0.5cm}
    \centering
    \includegraphics[width=\linewidth]{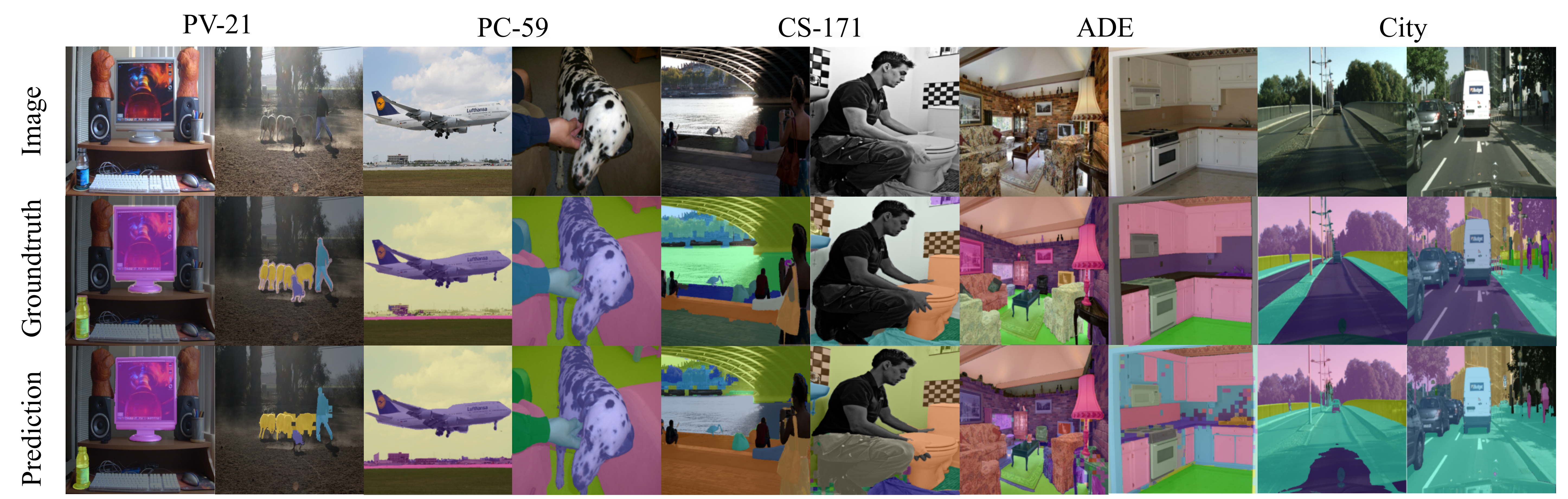}
    \vspace{-0.7cm}
    \caption{\small \textbf{Qualitative results of zeroshot segmentation.} The first row shows the ground truth labels. The second row shows the results of \ours-Stage 2 (refined).}
    \label{fig:pixelwise_seg}
\end{figure*}
\\
\noindent\myparagraph{Datasets.}
\textbf{COCO-Stuff~\cite{caesar2018coco}} contains 80 \textit{things} categories and 91 \textit{stuff} classes. We evaluate on both \textit{things only} (CO-80) and \textit{things + stuff} (CS-171) categories. 
 \textbf{Pascal VOC~\cite{everingham2010pascal}}   contains 20 foreground classes and everything else is labeled as background. To isolate the effect of the background class we evaluate on both (PV-20) excluding the background class and (PV-21) with the background class. For PV-21, we add \textit{stuff} classes from CS-171 as background categories in the evaluation of our method where these categories are mapped to the main background class. \textbf{Pascal Context~\cite{mottaghi2014role}}~(PC-59) contains 59 classes of objects and stuff. \textbf{CityScapes} ~\cite{cordts2016cityscapes} (City) contains 30 categories from street view images. \textbf{ADE20K}~\cite{zhou2019semantic} (ADE) contains 150 categories from various indoor and outdoor scenes. This dataset has the least overlap in terms of categories with our training dataset COCO and depicts the zeroshot performance of our method.\\  
 \begin{table*}[t]
    \begin{center}
    \footnotesize
   
    \resizebox{0.95\textwidth}{!}{
    \begin{tabular}{c|c|c|c|c|c|c|c|c}
    \toprule
    {Method}  &
{Training Data}& \multicolumn{7}{c}{mIoU} 
    \\
& &PV-21&PV-20& PC-59 &CO-80 &CS-171&City&ADE
  \\
   \midrule
    GroupVit~\cite{xu2022groupvit}&41M +Labels& 50.4&79.7&18.7 &27.5&15.3&11.1&9.2
   \\
                      Mask CLIP~\cite{zhou2022extract} &-&38.8 &74.9& 23.6&20.6 &16.4&12.6&9.8
   \\
   ReCo~\cite{shin2022reco}&-&25.1&57.7 & 19.9&31.6 &14.8&21.1&11.2
   \\
TCL~\cite{cha2023learning}&15M + Labels*&55.0&83.2 &33.9& 31.6 & 22.4&24.0&\underline{\textbf{17.1}}
\\
   \midrule
         ZeroSeg~\cite{chen2023exploring}& 3.4M*&-& 37.3&19.7&-&17.8&-&-
 \\
      CLIP-S4~\cite{he2023clip}& 0.12M*&  -& - &33.6&-&22.1&-&-
   \\
SCCLIP~\cite{wang2023sclip} & -&59.1&- &30.4& -& 22.4&-&-
\\
CLIP-DIY~\cite{wysoczanska2024clip}&-&59.0& -& 30.4&-&-& -&-
\\
CaR~\cite{sun2023clip}&-&67.6 &-&30.5 & 36.6&-&-&
\\
SAM-CLIP~\cite{wang2023sam}& 41M & 60.6 &-& 29.2& - &\underline{\textbf{31.5}}&-&\underline{\textbf{17.1}}
\\
   \midrule
                \ours - Stage 2 (base) & 0.12M (Stage 1 Annotations)*& \textbf{67.73}& \textbf{85.65}& \textbf{42.72}& \textbf{57.63}&29.88& \textbf{28.37}&16.25
                \\
                \ours- Stage 2 (refined) & 0.12M (Stage 1 Annotations)* &\underline{ \textbf{71.02}} &\underline{ \textbf{87.03}} &\underline{ \textbf{44.34}} &\underline{ \textbf{58.20}} &  \textbf{30.65} &\underline{ \textbf{30.55}} &  \textbf{16.60}
                \\

    \bottomrule
    \end{tabular}}
  
    \end{center}
    \vspace{-0.5cm}
       \caption{\small Semantic Segmentation performance on various datasets. Best method underlined, best and second best   marked in bold,   our method is better or on par with SOTA methods. Methods using COCO dataset as part of their training are marked by *.  \label{tab:segmentation-datasets}}
\end{table*}
\myparagraph{Compared Methods.}
We compare with methods that target the alignment of image features with language for open vocabulary semantic segmentation. We divide methods according to the annotation and training they require.
\textbf{Image Level Annotation:} 
GroupVit~\cite{xu2022groupvit} and TCL~\cite{cha2023learning}. \textbf{Annotation-free methods:}
We consider ZeroSeg \cite{chen2023exploring}, Mask CLIP~\cite{zhou2022extract} (the best performing variant), CLIP-S4~\cite{he2023clip} and  SAM-CLIP~\cite{wang2023sam}. 
\textbf{Training free methods:} 
CLIP-DIY~\cite{wysoczanska2024clip}, SCCLIP~\cite{wang2023sclip}, CaR~\cite{sun2023clip} and ReCo~\cite{shin2022reco}.

\myparagraph{Metrics.}
We consider the widely adopted Mean Intersection over Union (mIoU). We follow the evaluation protocol of TCL~\cite{cha2023learning}. We don't apply any post-refinement and only  one standard template \{a photo of a ``$class$''\} for text descriptions is considered unlike other approaches using 80 different templates during evaluation.
\subsection{Annotation-Free Semantic Segmentation}
\vspace{-0.1cm}
Table~\ref{tab:segmentation-datasets} reports the mIoU on different datasets. Results of methods in first block are taken from TCL~\cite{cha2023learning} while second block methods are taken from their corresponding papers. The third block reports our Stage 2 results when trained with the pseudo annotations generated on COCO-Stuff dataset.
\\
First,  training free methods, CLIP-DIY~\cite{wysoczanska2024clip},  SCCLIP~\cite{wang2023sclip}, CaR~\cite{sun2023clip} and ReCo~\cite{shin2022reco} perform inferior to our method mostly with a large margin.
Second, our method performs the best or second best on all datasets even improving over TCL~\cite{cha2023learning} trained on 15M images with image level labels. \ours-Stage 2 (base) outperforms TCL by a large margin of 8\% on PC-59 and 26\% on CO-80. Third, \ours-Stage 2 (base) outperforms SAM-CLIP~\cite{wang2023sam} significantly on PV-21 and PC-59 while being inferior on CS-171 and ADE albeit with a small margin. SAM-CLIP~\cite{wang2023sam} trains SAM encoder on 41M images with no code or model available which makes it infeasible for us to evaluate it on the remaining datasets. Nevertheless, we emphasize that our Stage 1 method is training free and our alignment module is light and trained only for a few epochs on almost 400 times less data compared to SAM-CLIP~\cite{wang2023sam}. This makes our method a plug and play approach  to semantically segment any dataset. Finally, adding refinement to FMbSeg pixel segmentation brings an average improvement of $1.7\%$.


\subsection{Qualitative Results}
First we inspect the alignment of patch features with vocabulary after we train our alignment module. Fig.~\ref{fig:alignment_heatmaps} shows the patch level similarity between the image features and the image level text features on Pascal VOC. The newly aligned image encoder produces distinctly more consistent similarity heatmaps for the given categories than those produced by CLIP. While the most lit patches in CLIP's heatmaps typically correspond to the queried object, CLIP's patch features are not sufficient to generate semantic segmentation masks of the objects. This shows the efficacy of aligning a better vision encoder with text rather than trying to improve CLIP patches features which typically requires large scale (re)training.
 Fig.~\ref{fig:pixelwise_seg} shows qualitative results of the annotation-free zeroshot segmentation  by \ours-Stage 2 (refined) on different datasets.

\subsection{Ablation}
\subsubsection{Stage 1 Ablation}
Table \ref{tab:stage1.1} evaluates the performance for the training-free part of our method, \ours-Stage 1. The generated pseudo annotations for each dataset are evaluated against the dataset's groundtruth segmentation masks. For ablation purposes only, we further provide the results for a semi-supervised variation of Stage 1.1 when the object presence detection part of the method (see \ref{subsec-1.1})  is replaced with the groundtruth image level annotations from the dataset. 

Stage 1.1 performs the lowest due the limitations mentioned
earlier, namely, small objects, single instance segmentations and wrong detections
due to visual/textual ambiguities (fig \ref{fig:Stag1}). Stage 1.2 achieves a relatively better performance compared to Stage 1.1 as it addresses
the limitations with small objects and can segment multiple instances. Semi-supervised Stage 1.1 removes the wrongly detected objects from the pipeline to vastly improve the performance, achieving comparable results to TCL\cite{cha2023learning}, a
state-of-the-art training-based method (Table~\ref{tab:segmentation-datasets}). These results demonstrate the effectiveness of our loss function design in overcoming the missed and incorrect predictions from Stage 1, leading to significantly improved performance. In \suppl we show that both Stage 1.1 and Stage 1.2 are essential for training the alignment module.

 \begin{table*}[t]
    \begin{center}
    \footnotesize
    \resizebox{0.9\textwidth}{!}{
    \begin{tabular}{l|l|l|l|l|l|l|l|l}
    \toprule
    {Method}  &
{Annotation}& \multicolumn{7}{c}{mIoU} 
    \\
& &PV-21&PV-20& PC-59 &CO-80 &CS-171&City&ADE
  \\
                \ours - Stage 1.1&- &25.37 & 42.92&22.03&32.97 &16.96&9.50&10.45
                \\
                \ours - Stage 1.2 &- & 36.68 & 53.36&24.25&29.15 &16.82&\textbf{14.86}&12.30
                \\
                \midrule
                \ours - Stage 1.1 (Semi-Supervised) & Image-level & \textbf{50.86}& \textbf{63.82}&\textbf{39.33}&\textbf{47.77} &\textbf{29.03}&{10.76}&\textbf{22.86}
                \\

    \bottomrule
    \end{tabular}}
  
    \end{center}
    \vspace{-0.5cm}
       \caption{\small  \textbf{Stage 1 ablation}. Stage 1.2 achieves a better performance as it can segment small objects and multiple instances of the same object. Semi-supervised Stage 1.1 employs image-level labels for mask generation and achieves a comparable performance to SOTA training based methods.\label{tab:stage1.1}}
\end{table*}

\vspace{-0.2cm}
\subsubsection{Architecture Ablation}
\vspace{-0.2cm}
We design our alignment module as a single transformer block with multi-head self-attention over image patches. We ablate our choice against other designs, namely a single linear layer and a Multi-layer perceptron (MLP) with GELU activation. 
Table~\ref{tab:loss_arch_ablation} reports the mIOU on CS-171. The differences are not substantial, with MLP achieving 
better performance than linear. The transformer block further improves over the MLP.

\begin{table}[h!]
  \centering
   \begin{subtable}[h]{0.49\linewidth}
        \centering
    \resizebox{\linewidth}{!}{
     \begin{tabular}{l|l}
\toprule
\footnotesize
      Architecture & mIoU \\
\midrule
     Linear &27.25\\
     MLP & 28.90\\
     Transformer block & 29.88\\
\bottomrule
\end{tabular}}
   \end{subtable}
    \hfill
   \begin{subtable}[h]{0.49\linewidth}
        \centering
     \resizebox{\linewidth}{!}{
        \begin{tabular}{l|l}
    \toprule
   Alignment Loss  &
   { mIoU }
  
    \\

        \midrule    
        $\ell_{SupCon}$&  26.25\\
       $\ell_{\pt}$ (2) & 28.51\\
       $\ell_{\ESupCon}$ (1)& 29.88
\\
    \bottomrule
    \end{tabular}}
     \end{subtable}
      \caption{ \small CS-171 mIoU with our base model. Left. \textbf{Comparison of different design choices for our alignment module}, a transformer block has a small advantage. Right. \textbf{Comparison of different losses}. Our full loss \ESupCon performs the best. \label{tab:loss_arch_ablation}}
      \vspace{-0.1cm}
    \end{table}
\subsubsection{Loss Ablation}\label{sec:loss-ablation}
\vspace{-0.2cm}
In section \ref{subsec-alignmentmodule} we introduced our loss function for aligning patch features by contrasting them with each other and with text features. Here, we compare to two  loss variants:
1) The supervised contrastive loss~\cite{khosla2020supervised} (SupCon) alone where text features are considered as examples of the corresponding concepts similar to the image patches of a given category. 2) The prototype loss alone, Eq.\ref{eq:l_clSup}. 
Table~\ref{tab:loss_arch_ablation} reports the mIOU on CS-171. We find that SupCon term alone exhibits the worst convergence while the prototype loss shows a stronger performance, probably due to the  smooth approximation of Cross Entropy loss and SupCon loss on text-patch pairs. However, when combined with SupCon on pairs of patches, better performance is achieved due to more enhancement in the expressivity of  these patches features. 
Overall, the unique treatment of \ESupCon to the text features allows a smoother generalization, better convergence and hence considerably better performance.
\subsection{Annotation-Free Segmentation Applications}
To further illustrate the advantages of our open vocabulary segmentation tool, in this section we evaluate our annotation free semantic segmentation method on tasks different from those evaluated by common segmentation benchmarks. These serve as examples for different scenarios where it is required to segment a specific object with no pixel-level training data/pretrained model available. 
\vspace{-0.5cm}
\subsubsection{Plug and Play Binary Segmentation Task}
We consider a water segmentation task based on WaterV2 dataset from Kaggle\footnote{https://www.kaggle.com/datasets/gvclsu/water-segmentation-dataset}. Water's uniform appearance, lighting conditions and anomalies (i.e. objects or reflections inside the water) make this a difficult segmentation task. Fig \ref{fig:supp_water_seg} shows some qualitative results for the water segmentation task. \ours Stage 1 achieves an mIoU accuracy of $83.1\%$ on the evaluation set of this dataset.
\subsubsection{LLM Personalization Task}
Here we briefly showcase our method's ability to fit into a completely different task. The main goal  is to identify specific instances of objects such as personal items (My cat, my running shoes, my espresso cup etc.) given very limited number ($=4$) training views for each object. Figure \ref{fig:personalization} illustrates a few examples of the personalized objects in the training and evaluation set.\\
With no training, we employ our Stage 1 method to segment out each object in the training set given its name.
For the test images, we query the model to generate a mask for each personalized objects on every image. we measure the cosine similarity of the DINOv2 features corresponding to the segmented mask for the personalized object with those extracted from the corresponding training images. A high similarity would indicate the presence of the object instance.\\
We compare our simple pipeline to a training based method where a new classification head is trained for each personalized object instances~\cite{alaluf2024myvlm}. As seen in table \ref{tab:personalization}, by just using a similarity threshold for detection, our stage 1  performs comparable to \cite{alaluf2024myvlm} on 29 personalized objects without any modification for this task.

\begin{figure}
    \centering
    \includegraphics[width=\linewidth]
    {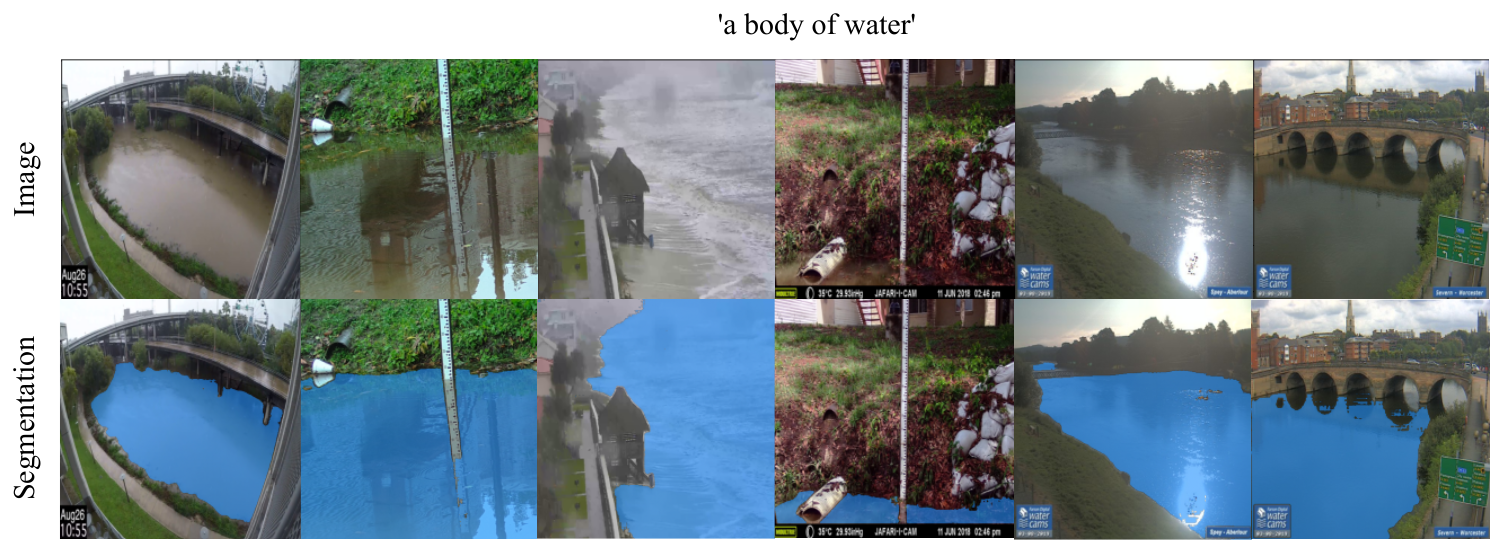}
\vspace{-0.6cm}    \caption{\textbf{Water segmentation results:} Stage 1 accurately segments bodies of water in presence of anomalies and different lighting conditions achieving a $83.1\%$ mIoU accuracy.}
\vspace{-0.5cm}
    \label{fig:supp_water_seg}
\end{figure}

\begin{figure}
    \centering
\includegraphics[width=0.9\linewidth]{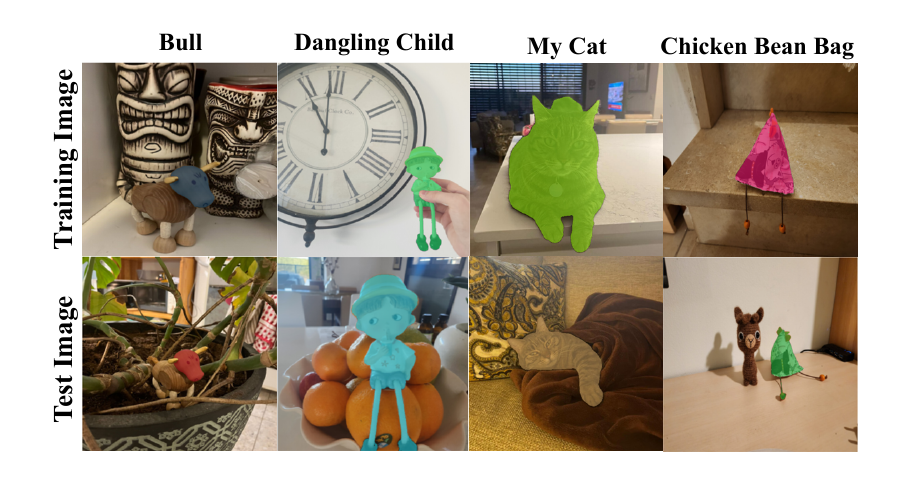}
    \vspace{-0.4cm}
    \caption{\small \textbf{Personalized object segmentation.} Our Stage 1 method can segment out the personalized items with no training.}
    \label{fig:personalization}
\end{figure}

 \begin{table}
    \begin{center}
    \resizebox{0.6\linewidth}{!}{
    \begin{tabular}{l|l|l}
\toprule
\footnotesize
      Method & Precision & Recall \\
\midrule
     MyVLM \cite{alaluf2024myvlm} & 90\% & 96\%\\
     \ours-Stage 1 & 91\% & 87\%\\
\bottomrule
\end{tabular}}
  
    \end{center}
    \vspace{-0.5cm}
       \caption{\small \textbf{Personalized object retrieval.} Our out of the box method performs comparable to MyVLM's classification heads trained specifically for this task.  
       \label{tab:personalization}}
   \vspace{-0.5cm}
\end{table}

\label{sec-experiments}
\vspace{-0.5cm}
\section{Conclusion and Future Work}
\label{sec-conclusion}
Given the impressive performance of vision-language foundation models and their transferability to various downstream tasks, we consider the problem of open vocabulary annotation-free semantic segmentation. By composition of different foundation models, namely CLIP and SAM, we extract free pixel level annotations. We then propose a lightweight alignment module that projects the embedding of any arbitrary pretrained vision encoder to the text encoder space. Our method  can be deployed as a plug-and-play customized alignment module for any semantic segmentation dataset with zero annotations. We show SOTA results demonstrate the effectiveness of foundational models compositionality. As future work, we want to investigate other image encoders and  continuous fine-tuning on new categories. Additionally, we want to explore how our alignment module can improve VLM models that are based on CLIP to further strengthen their object localization capabilities.

{
    \small
    \bibliographystyle{ieee_fullname}
    \bibliography{egbib}
}
\end{document}


\title{Appendix: Annotation-Free Semantic Segmentation with Vision Foundation Models}

\maketitle

\section{Stage 1}
\label{sec:method_details}
In this section we further explain  our design choices for pseudo label generation with the training free part of our method.
\subsection{Stage 1.1}
\subsubsection{Patch-level Feature Extraction}
CLIP model has been pretrained on image/text pairs to provide an \textit{image-level} classification of an input given the candidate text queries. In this work, we employ CLIP to extract \textit{patch-level} similarities of image/text pairs.

\begin{figure*}
    \centering
    \includegraphics[width=0.8\linewidth, height=9.5cm]{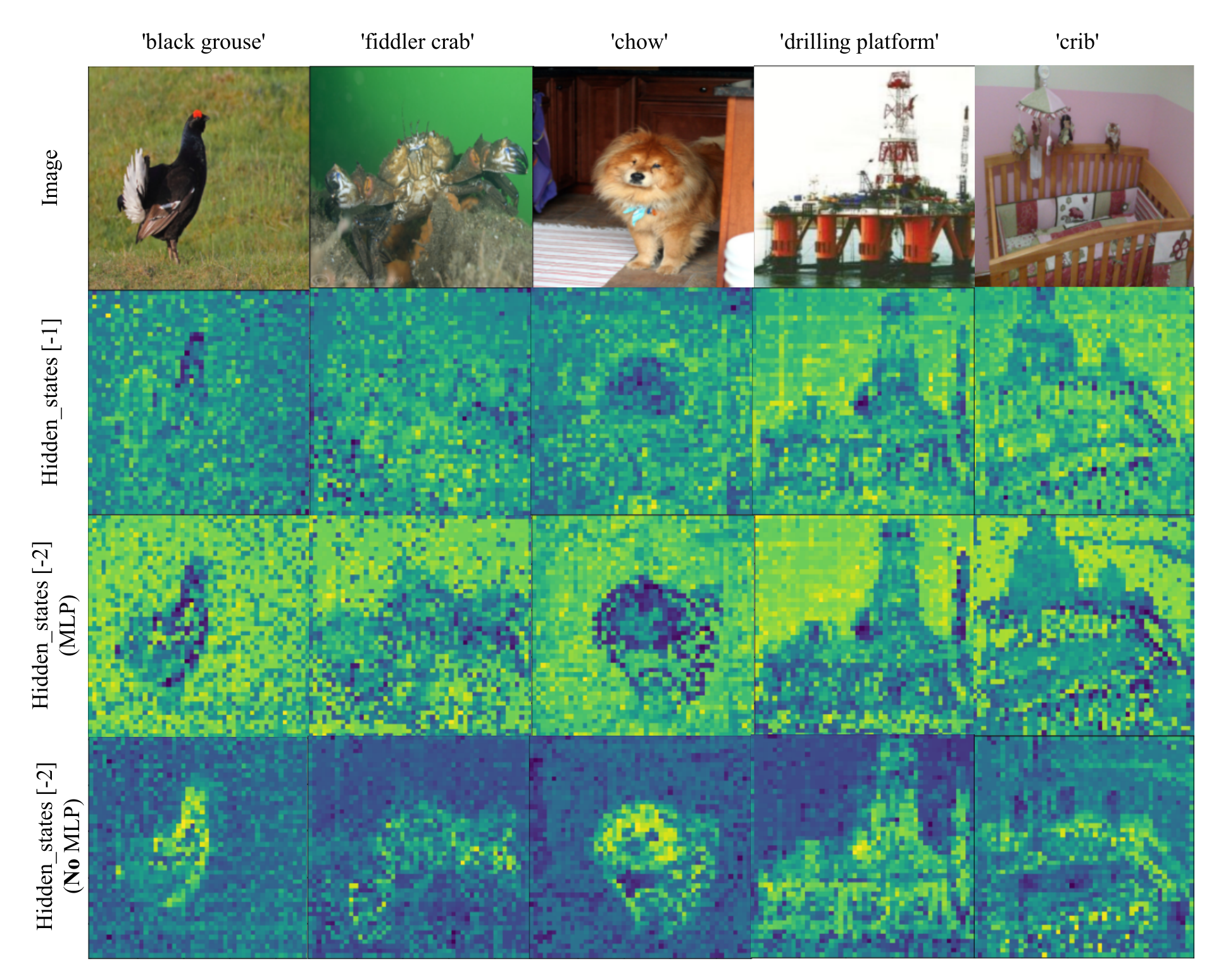}
\vspace{-0.4cm}    \caption{\textbf{Patch-level feature extraction:} CLIP vision encoder's last hidden state shows negative alignment with object's text embeddings. We alleviate this by skipping the average pooling, self-attention \cite{zhou2022extract} and the MLP in the last layer of CLIP's architecture. Heatmaps are generated at original image resolution of $672\times672$ for this figure.}
    \label{fig:supp_inversion}
\end{figure*}

The most straightforward approach to extract patch-level features from CLIP's vision encoder is to access the last hidden state and pass it through CLIP's visual projection layer. However, we observe a negative alignment between the text/patch embeddings. Patches representing the object typically show the least similarity to the corresponding text query of the image label (figure \ref{fig:supp_inversion}, row 2).  

Instead, we follow a similar procedure introduced in \cite{zhou2022extract} to extract patch embeddings. We notice that the final MLP layer in the architecture causes a negative alignment of the text/patch features (figure \ref{fig:supp_inversion}, row 3). Therefore, we remove this layer from the network which results in a correct alignment of the text/patch similarities (figure \ref{fig:supp_inversion}, row 4).

\subsubsection{High Resolution Heatmap Generation}
\label{sec:highres_heatmaps}
\begin{figure*}
    \centering
    \includegraphics[width=0.8\linewidth, height=10cm]{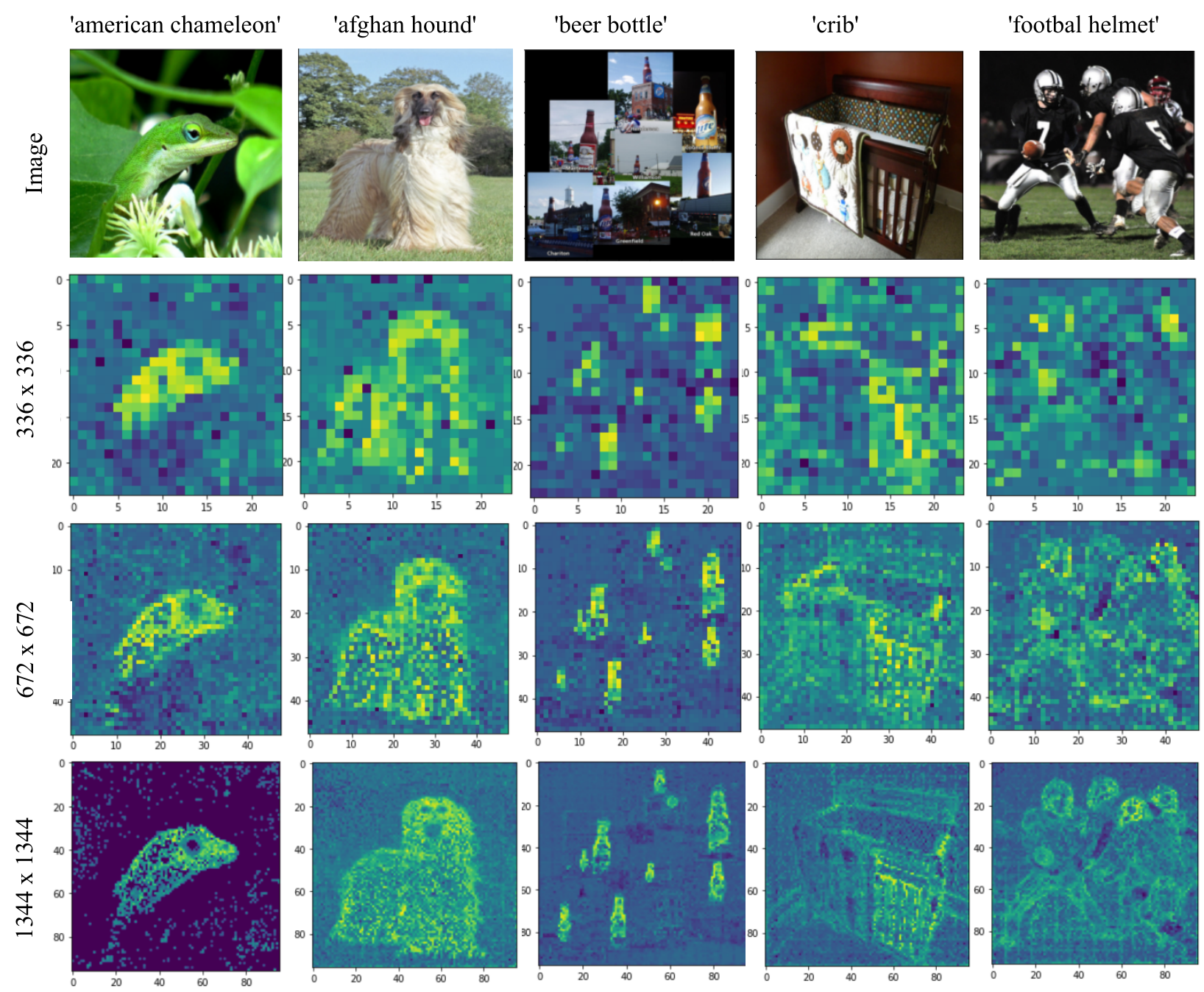}
    \vspace{-0.4cm} 
    \caption{\textbf{High-resolution heatmap/feature generation:} We generate more precise heatmaps by oversampling and processing image crops separately. Heatmaps are generated on examples from ImageNet-1k dataset \cite{ILSVRC15}.}
    \label{fig:supp_res}
\end{figure*}

Our CLIP Vit-L/14 model accepts inputs of size $336\times 336$. With such resolution, our model roughly localizes objects in the images (for Stage 1.1) and the feature maps are sub-optimal for labelling SAM masks (Stage 1.2) (figure  \ref{fig:supp_res}, row 2). Therefore, more precise feature maps can directly boost SAM's performance for our Stage 1 method. 

We achieve this by oversampling the images. Particularly, we divide the oversampled image into non-overlapping crops of CLIP's input size (i.e. $336\times 336$). Patch features for each crop is generated by CLIP's vision encoder and features for all crops are gathered into one single feature map representing the high-resolution image (figure \ref{fig:supp_res}, row 3 and 4).

Table \ref{tab:supp_crops} details the co-relation between the image resolution, number of crops per image and the final feature map size. Although a higher resolution image results in a more precise feature map, it would require a higher computation. To keep a trade off, we generate our feature maps with an input resolution of $1344\times1344$ and 16 total number of crops per image for all experiments in the paper.
\subsubsection{Label-free Object Segmentation}
\label{sec:label_free}
\vspace{-0.2cm}
As mentioned in section 3.2 of the main paper, the $\cls$ token for each crop is used to classify it based on its similarity to the text features of classes in $V$. In case an object class was assigned to more than a threshold ($\mathcal{T}$) number of crops, we mark the class as \textit{detected} and the method proceeds to generate a segmentation mask for the object in the input image (See section \ref{sec:sam_details} for more details). Otherwise, the class is discarded. However, if an image-level label is present, the method proceeds to generate the mask without a threshold check. We refer to this as the \textit{semi-supervised} variation of our method in the main paper. We set ($\mathcal{T} = 1$) for all the label-free experiments in the paper.

\vspace{-0.5cm}   

\begin{table}[t]
    \begin{center}
    \footnotesize
    \resizebox{0.49\textwidth}{!}{
       \begin{tabular}{l|l|l|l}
    \toprule
    Resolution &
    Number of Crops &
  Patch Embedding Size
  
    \\

        \midrule
            $336\times336$  & 1 & $24\times24\times768$
    \\
        \midrule 
        $672\times672$  & 4 & $48\times48\times768$\ 
\\
    \midrule
    \textcolor{green}{$1344\times1344$} & \textcolor{green}{16} & \textcolor{green}{$96\times96\times768$}\
\\
    \bottomrule
    \end{tabular}}
    \caption{\textbf{Resolution setting} for generating image feature maps using CLIP Vit-L/14. We oversample the images to a higher resolution and generate more precise feature maps. For experiments in the main paper we generate feature maps with the setting marked by green in the table.}
    \label{tab:supp_crops}
    \vspace{-0.8cm}
    \end{center}
\end{table}

\begin{figure*}
    \centering
    \includegraphics[width=\linewidth]{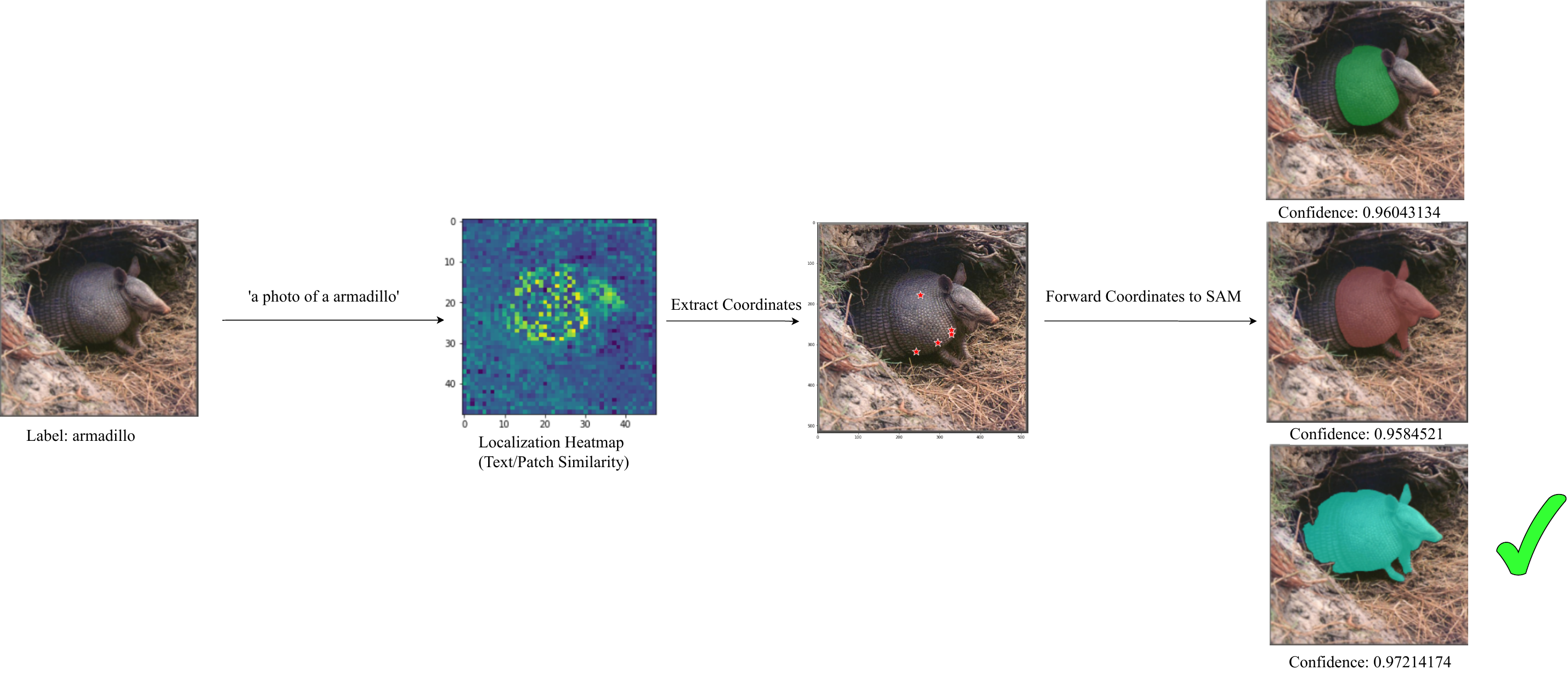}
    \caption{\textbf{Query point selection and mask generation with SAM:} Our method selects 5 patches with the highest similarity to object's text embedding. The coordinates for the center of these patches are forwarded to SAM which generates 3 different segmentation masks for the object. Our method selects the segmentation mask with the highest confidence. Input image is taken from ImageNet-1k dataset.}
    \label{fig:supp_sam}
\end{figure*}

 \begin{table*}[h!]
    \begin{center}
    \footnotesize
    \resizebox{0.95\textwidth}{!}{
       \begin{tabular}{l|l|l|l|l|l|l|l|l|l|l}
    \toprule
     \multirow{2}[1]{*}{Pseudo Annotations}  &
 \multirow{2}[1]{*}{Stage 1.1 Split}& \multirow{2}[1]{*}{Stage 1.2 Split }& \multicolumn{7}{c}{mIoU} 
    \\
& &&PV-21&PV-20& PC-59 &CO80 &CS-171&City&ADE&AVG.
  \\
      \midrule
                Stage 1.1 + 1.2 & 100\%& 0\%  &  65.83&85.41&41.34&57.56 &28.02&22.15&15.40&45.10
                \\
                Stage 1.1 + 1.2 & 100\%& 33\%  & 
                \textbf{67.89}&84.92&\textbf{43.06}&57.19&28.85&26.71&16.11&46.39
                \\
                Stage 1.1 + 1.2 & 100\%& 75\%  & 67.73&\textbf{85.65}&{42.72}&\textbf{57.63}&\textbf{29.88}&\textbf{28.37}&\textbf{16.25}&\textbf{46.89}
                \\
                Stage 1.1 + 1.2 & 100\%& 100\%  & 
                63.58&83.90&38.69&54.99&27.55&28.45&14.99&44.59
                \\

    \bottomrule
    \end{tabular}}
    \caption{\textbf{Stage 1.2 Data Balancing Ablation}. Since Stage 1.2 generates many more pseudo annotations than Stage 1.1, there is a need to balance the training data for training the alignment module. For all experiments we use a randomly sampled 75\% split from Stage 1.2.}
    \label{tab:stages_balance}
    \end{center}
\end{table*}

\subsubsection{SAM Details}
\label{sec:sam_details}
We detail our design choices involving SAM in this section. Figure \ref{fig:supp_sam} summarises the steps for generating a segmentation mask for an image given its label. We follow the same procedure for all the \textit{detected} objects (section \ref{sec:label_free}) in case the image-level label is not available.

\noindent\textbf{Resolution:} SAM supports segmentation on any input resolution. Since we work on top of a DINOv2 model for the Stage 2 of our method (alignment module) , we select the input size for SAM to be the same as DINOv2 model, namely $518\times518$.

\noindent\textbf{SAM query points:} As mentioned earlier, we generate our localization heatmaps for images of size $1344\times1344$. This would result in a patch-level localization of $96\times96$ (table \ref{tab:supp_crops}). We take 5 highest activated patches in the localization heatmap for each detected object. Such patches have the most similar visual embedding to the text embedding of the detected object. Next, our method converts the coordinates for the center of those patches to coordinates in $518\times518$. These points are forwarded to SAM along with the image to generate the segmentation mask for the corresponding object.

\noindent\textbf{Confidence based segmentation mask selection:} SAM generates 3 masks for each set of query points to account for ambiguity. Besides, it produces a confidence measure (estimated IoU) for each mask. We select the mask with the highest confidence as our final segmentation mask for the detected object.
\subsection{Stage 1.2}
Here we briefly detail stage 1.2 design choices and provide visual examples to complement Figure 3 in the main paper.
\subsubsection{Qualitative Evaluation}
\begin{figure*}
    \centering
    \includegraphics[width=0.9\linewidth]{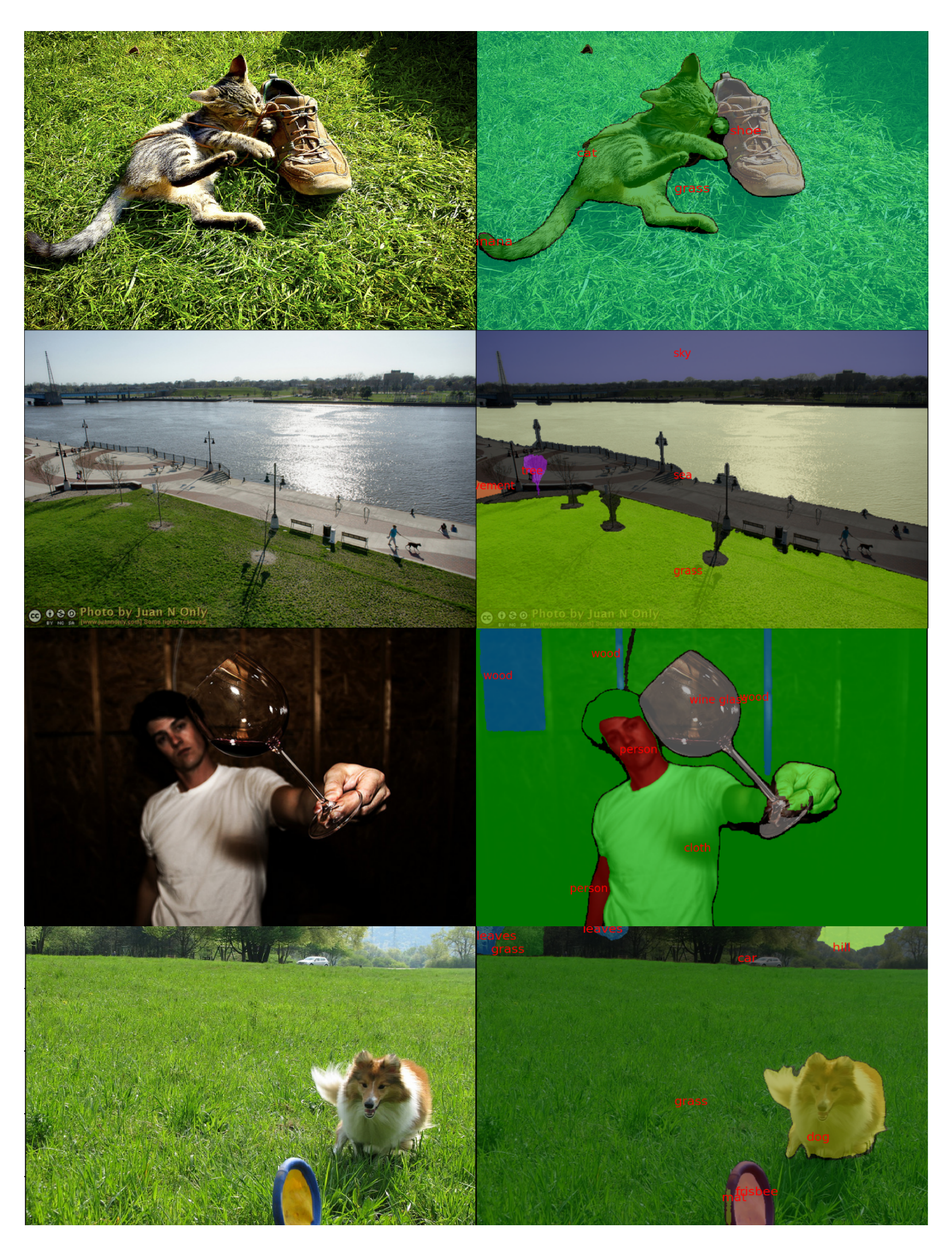}
    \caption{\textbf{Qualitative Evaluation of Stage 1.2}. Segmented objects and their corresponding labels (in red) are shown on the image. SAM's automatically-generated masks come without a semantic label. Our method assigns the label as the class with the highest similarity of its text features to CLIP's mean embedding of the mask. This labelling strategy might occasionally fail. However, Stage 1.2 proves to be a complementary method to adress Stage 1.1 limitations. Images are taken from COCO dataset.}
    \label{fig:stage_1.2}
\end{figure*}

Figure \ref{fig:stage_1.2} shows examples of images segmented by Stage 1.2. We employ SAM to generate non-semantical object segmentations. Particularly, we initialize SamAutomaticMaskGenerator object class with $iou\_pred\_threshold=0.97$. This is a measure to filter out masks with a low quality. Such a high value removes many of the overlapping and redundant masks. However, it might also exclude some regions of the images from the mask generation (road/pavement in the second image). 
\vspace{-0.05cm}
For each mask we crop the patch-features corresponding to the area covered by the mask and calculate the average CLIP embedding for that region. Next, the class with the highest text feature similarity to this mean embedding is selected as the mask's semantic label. While Stage 1.2 might occasionally fail in assigning the correct label to each mask, it alleviates the issue with small objects and partial masks generated by Stage 1.1.

\section{Stage 2}
Here we detail our design choices for the training based part of our method.
\subsection{Data Balancing Ablation}
Since Stage 1.2 annotates all the possible masks in each image, the number of annotations produced by Stage 1.2 is generally much higher than Stage 1.1. This might cause an imbalance when training the alignment module with both Stage 1.1 and Stage 1.2 pseudo annotations. Table \ref{tab:stages_balance} demonstrates the performance of Stage 2 (alignment module) when trained with different ratios of randomly sampled pseudo annotations from Stage 1.2. As demonstrated in this table, having the full annotations from Stage 1.2 (row 4) performs even worse than training only with Stage 1.1 annotations (row 1). Therefore, in order to avoid Stage 1.2 overpowering the training, in our experiments we trained the alignment module with a 75\% split of Stage 1.2 pseudo annotations (row 3).

{
    \small
    \bibliographystyle{ieee_fullname}
    \bibliography{egbib}
}